\documentclass[11pt]{article}

\usepackage[final]{acl}

\usepackage{times}
\usepackage{latexsym}
\usepackage[table]{xcolor}

\usepackage[T1]{fontenc}

\usepackage[utf8]{inputenc}

\usepackage{microtype}

\usepackage{inconsolata}

\usepackage{graphicx}

\title{Developing an English–Efik Corpus and Machine Translation System for Digitization Inclusion}

\author{
 \textbf{Offiong Bassey Edet\textsuperscript{1,2}},
 \textbf{Mbuotidem Sunday Awak\textsuperscript{2}},
 \textbf{Emmanuel Oyo-Ita\textsuperscript{1}},
\\
 \textbf{Benjamin Okon Nyong\textsuperscript{3}},
 \textbf{Ita Etim Bassey\textsuperscript{4}},
\\
\\
 \textsuperscript{1}University of Cross River State,
 \textsuperscript{2}ML Collective,
 \textsuperscript{3}Arthur Jarvis University,\\
  \textsuperscript{4}University of Calabar
\\
\small{
\href{mailto:offiongbassey99@gmail.com}{offiongbassey99@gmail.com}, 
\href{mailto:mbuotidemawak@gmail.com}{mbuotidemawak@gmail.com}, 
\href{mailto:emmanueloyoita@unicross.edu.ng}{emmanueloyoita@unicross.edu.ng},
}\\
\small{
\href{mailto:benokon26@gmail.com}{benokon26@gmail.com},
\href{mailto:basseyita141@gmail.com}{basseyita141@gmail.com}
}
}

\begin{document}
\maketitle
\begin{abstract}
Low-resource languages serve as invaluable repositories of human history, preserving cultural and intellectual diversity. Despite their significance, they remain largely absent from modern natural language processing systems. While progress has been made for widely spoken African languages such as Swahili, Yoruba, and Amharic, smaller indigenous languages like Efik continue to be underrepresented in machine translation research. This study evaluates the effectiveness of state-of-the-art multilingual neural machine translation models for English–Efik translation, leveraging a small-scale, community-curated parallel corpus of 13,865 sentence pairs. We fine-tuned both the mT5 multilingual model and the NLLB-200 model on this dataset. NLLB-200 outperformed mT5, achieving BLEU scores of 26.64 for English–Efik and 31.21 for Efik–English, with corresponding chrF scores of 51.04 and 47.92, indicating improved fluency and semantic fidelity. Our findings demonstrate the feasibility of developing practical machine translation tools for low-resource languages and highlight the importance of inclusive data practices and culturally grounded evaluation in advancing equitable NLP.
\end{abstract}

\section{Introduction}
The rapid advancement in Artificial Intelligence (AI), especially in the field of Natural Language Processing (NLP), has significantly enhanced the ability to translate languages accurately \cite{Vaswani2017Attention, Fan2021BeyondEnglsih}. However, not all languages have equally benefited from these technological improvements. Many indigenous languages, particularly those classified as “low-resource,” lack sufficient linguistic datasets and digital tools needed for developing effective translation systems \cite{Joshi2020TheState, Nekoto2020Participatory}. One such language is Efik, spoken primarily in the southern part of Cross River State, Nigeria; particularly Calabar South, Calabar Municipality, Odukpani, Akpabuyo, Bakassi and parts of Akamkpa. Efik is also spoken in some parts of Akwa Ibom State in Nigeria and northwestern Cameroon \cite{Mensah2023Melting, Offiong2013EfikLanguage, Mensah2014TheAdaptation}. 

Effective translation technologies that bridge English and Efik could therefore have substantial socio-economic and cultural benefits, enhancing communication, fostering inclusion and preserving cultural identity.

Lack of data remains a critical problem in developing neural machine translation systems for African low-resource languages. Most available web data are small and of poor quality \cite{Adelani2022}. Existing systems often fail to translate idiomatic expressions, cultural nuances, and linguistically complex structures accurately, which affects overall translation quality. The primary challenge is the severe scarcity of parallel corpora (sets of sentences paired in English and Efik), which are essential for training AI translation models \cite{Agić2019}. Recent shared tasks such as the AmericasNLP 2025 challenge have shown that MT for Indigenous low-resource languages remains extremely difficult due to severe data scarcity and domain-specific linguistic complexity \cite{Gibert2025AmericaNLP}. 

Given these challenges, this study evaluates the effectiveness of state-of-the-art multilingual machine translation models, specifically mT5 and NLLB-200, fine-tuned for English-to-Efik translation.

\section{Overview of Efik}
Efik is the indigenous language of the Efik people, whose major settlements lie in the southern part of Cross River State in Southeastern Nigeria \cite{Mensah2023Melting}. Efik is one of the first languages in Nigeria to be written and systematically studied. Its early prestige led missionaries to develop the orthography, grammar, lexicon, and a substantial body of literature \cite{Offiong2013EfikLanguage}.

The language functions as a lingua franca in southern Cross River State and is widely used in radio and television programs, markets, churches, and political activities such as campaigns and rallies. It is also taught in most public primary and secondary schools in the region, as well as in the Linguistics departments of the University of Calabar and the Cross River University of Technology. Several church denominations, including The Apostolic Church Nigeria, Mount Zion Church, and The Presbyterian Church, use Efik during services, often alongside real-time interpretation into English.

Efik was the first Nigerian language to be reduced to writing \cite{Noah2004EfikOrthography} and the second to receive a full Bible translation after Yoruba \cite{Offiong2013EfikLanguage}.

\subsection{Decline and Revitalization Efforts}
\subsubsection{Decline}
Efik once featured prominently in national and regional broadcasting, including Radio Nigeria’s news programs in the late 1960s through the early 1980s. Today, however, it is no longer used in national broadcast media. The language was also dropped at one point by the West African Examinations Council (WAEC), despite its long history as a written language. Several factors contribute to its declining use: limited technological support, insufficient institutional encouragement and decreasing intergenerational transmission. Few new materials have been developed in recent decades compared with earlier periods. In addition, many families now prioritize English over Efik. In numerous households and neighborhoods, English is used in everyday communication, often due to the misconception that fluency in English is a marker of intelligence. As a result, many parents discourage children from learning or speaking Efik, a trend that threatens the vitality of the language in future generations \cite{Offiong2013EfikLanguage}.

\subsubsection{Revitalization Efforts}
Historically, various attempts have been made to expand Efik orthography and vocabulary, though many did not achieve widespread adoption. The long-standing reliance on English loanwords continues to create challenges in developing a robust, fully standardized writing system \cite{Offiong2013EfikLanguage}.

Community tools such as the online Tete Dictionary have emerged to support Efik literacy and vocabulary access for learners \cite{Odey2025EfikImmersion}.

\section{Related Works}
Earlier research by \citet{Arikpo2018EngEfik} focused on the development of an automated English–Efik machine translation system using a rule-based approach rather than modern Neural Machine Translation (NMT). The system relied on a small bilingual dictionary containing 500 English–Efik words and manually defined transfer rules to perform translations. While it produced reasonable results for simple sentences, its performance degraded significantly on complex or composite sentences, highlighting the limitations of rule-based methods in handling diverse syntactic structures, morphological inflections, and idiomatic expressions. The study represents an early attempt to automate translation for Efik but was constrained by the limited dictionary size, absence of robust evaluation metrics, and inability to generalize beyond the manually encoded rules. Despite these limitations, it remains a notable first effort in English–Efik machine translation and underscores the need for larger, data-driven corpora and modern NLP techniques to improve translation quality for low-resource languages.

The JW300 parallel corpus, first collected by \citet{Agić2019}, provided a large-scale set of English-Efik Bible verse pairs and became foundational for early translation research. Building on this, the Masakhane community initiative \cite{Nekoto2020Participatory} demonstrated the effectiveness of crowdsourced data: volunteers across Africa aligned Efik text, and their Transformer-based machine translation model achieved BLEU scores of 33.48 for English–Efik and 33.68 for Efik–English, demonstrating strong translation performance and contributing open evaluation protocols.

More recently, \citet{Kalejaiye2025IbomNLP} introduced the Ibom NLP dataset, covering several Akwa Ibom languages, including Efik. Their work demonstrates the effectiveness of fine-tuning massively multilingual models, such as M2M-100 and NLLB-200 \cite{NLLB2022}, for low-resource Nigerian languages. On English–Efik translation, the models achieved BLEU scores up to 36, while for Efik–English the best BLEU score reached 34.6, highlighting the potential of cross-lingual transfer even with limited data.

While previous works such as \citet{Nekoto2020Participatory} report higher BLEU scores on JW300-based English-Efik data, these results are largely domain-specific (religion-bible). In contrast, our corpus introduces more diverse, natural data across multiple domains, including daily conversation, family and relationships, agriculture, environment, health, and education, emphasizing generalization rather than domain memorization.

\section{Methodology}
\subsection{Dataset Creation}
All English sentences were translated by six native Efik speakers and two trained Efik linguists, and independently verified by two additional linguists. All annotators are fluent in both English and Efik. We initially translated 14,000 English sentences into Efik. After removing duplicate entries, incomplete translations, and noisy sentence pairs during data cleaning, the final dataset contains 13,865 high-quality English–Efik parallel pairs. The dataset is stored in UTF-8 encoded JSONL format for efficient preprocessing and reproducibility.

Each sentence was assigned to one primary translator, who produced a single Efik translation. The translation was then independently reviewed by at least two different annotators who did not produce the original translation. Where disagreements occurred, a linguist reviewer performed final adjudication and correction.

\subsubsection{Data Collection}
English source sentences were created through a combination of controlled large language model (LLM) generation across multiple domains, manually written sentences by Efik–English bilingual linguists and native speakers, and example sentences drawn from a printed Efik–English dictionary. The LLM prompts were designed to generate short, domain-diverse sentences covering areas such as daily conversation, food, narratives, religion, and general knowledge. Dictionary examples were manually selected, digitized, and verified to ensure correctness and contemporary usage.

Annotators were instructed to preserve semantic meaning while ensuring natural Efik phrasing rather than literal word-for-word translation. The Learner’s Dictionary of the Efik Language \cite{AyeEfikDictionary} was used as a reference resource to resolve lexical ambiguity, verify word meanings, and confirm orthographic conventions, particularly for less frequent or older vocabulary. Each translated sentence was reviewed by at least two independent annotators to ensure grammatical correctness, semantic fidelity, and naturalness.

\subsubsection{Domain Coverage}
Table ~\ref{tab:domain-coverage} shows the domain distribution of the collected English sentences before final cleaning.

The dataset covers a diverse range of domains, with the largest portion coming from daily conversation (7,245 sentences), followed by family and relationships (1,885) and food and agriculture (1,237). Smaller categories include nature and environment, religion, time and number, health, education, and greetings. Sentences that did not clearly fit into any of these categories were grouped under “others” (872), ensuring that all source material was accounted for while maintaining meaningful domain classification.

\begin{table}[ht]
  \centering
  \small
  \begin{tabular}{lc}
    \hline
    \textbf{Domain} & \textbf{Number of Sentences} \\
    \hline
    Daily Conversation       & 7,245  \\
    Family and Relationship  & 1,885  \\
    Food and Agriculture     & 1,237  \\
    Nature and Environment   & 855    \\
    Religion                 & 512    \\
    Time and Number          & 509    \\
    Health                   & 475    \\
    Education                & 289    \\
    Greetings                & 121    \\
    Others                   & 872    \\
    \hline
  \end{tabular}
  \caption{Domain distribution of sentences.}
  \label{tab:domain-coverage}
\end{table}

\subsubsection{Data Extraction}
Each parallel pair was stored as a JSON object using the following structure:

\begin{quote}
\{
    "eng": "my sister is brushing her hair before going to school.",  
  
    "efi": "eyeneka mi ke asari idet esie mbemiso enye aka uf\d{o}k\~nwed." 
\}
\end{quote}

Each line in the JSONL file represents one English–Efik sentence pair, making the dataset easy to preprocess and compatible with standard NLP toolkits such as HuggingFace Transformers, OpenNMT, and Fairseq.

\subsection{Data Preprocessing}
\subsubsection{Data Cleaning}
The dataset underwent cleaning steps including removal of duplicates, eliminating extra spaces and standardizing text encoding to UTF-8. Sentences with incomplete or inconsistent translations were discarded.

\subsubsection{Dataset Splitting}
After cleaning and normalization, the English–Efik parallel corpus containing 13,865 sentence pairs was split into 10,398 for training, 1,387 for validation, and 2,080 for testing, as summarized in Table ~\ref{tab:train-test}.

\begin{table}[ht]
  \centering
  \begin{tabular}{lc}
    \hline
    \textbf{Split} & \textbf{Number of Sentences} \\
    \hline
    \verb|Train|     & {10,398}           \\
    \verb|Dev|     & {1,387}           \\
    \verb|Test|     & {2,080}           \\
    \hline
    \textbf{Total}     & \textbf{13,865}           \\
    \hline
  \end{tabular}
  \caption{Dataset Split.}
  \label{tab:train-test}
\end{table}

\subsection{Dataset Statistics and Analysis}
\subsubsection{Qualitative Overview}
The dataset contains 13,865 English–Efik parallel entries, spanning a wide range of everyday communication.

It includes both sentence-level translations and word or phrase-level entries, particularly for culturally specific terms that require contextual explanation. Lexically, the corpus contains 11,080 unique English word types with 103,734 tokens, and 11,218 unique Efik word types with 100,458 tokens.

Both English and Efik share the same longest sentence in the dataset, consisting of 57 words, while the shortest entries are single-word translations. This distribution reflects a mixture of conversational exchanges, domain-general expressions, and longer narrative or didactic passages.

\subsubsection{Sentence Length Distribution}
English sentences have an average length of approximately 6 words, with most sentences falling between 2 and 12 words.
 Efik sentences have an average length of approximately 5 words, with most entries ranging between 2 and 10 words.
 
\begin{figure}[ht]
  \centering
  \includegraphics[width=\linewidth]{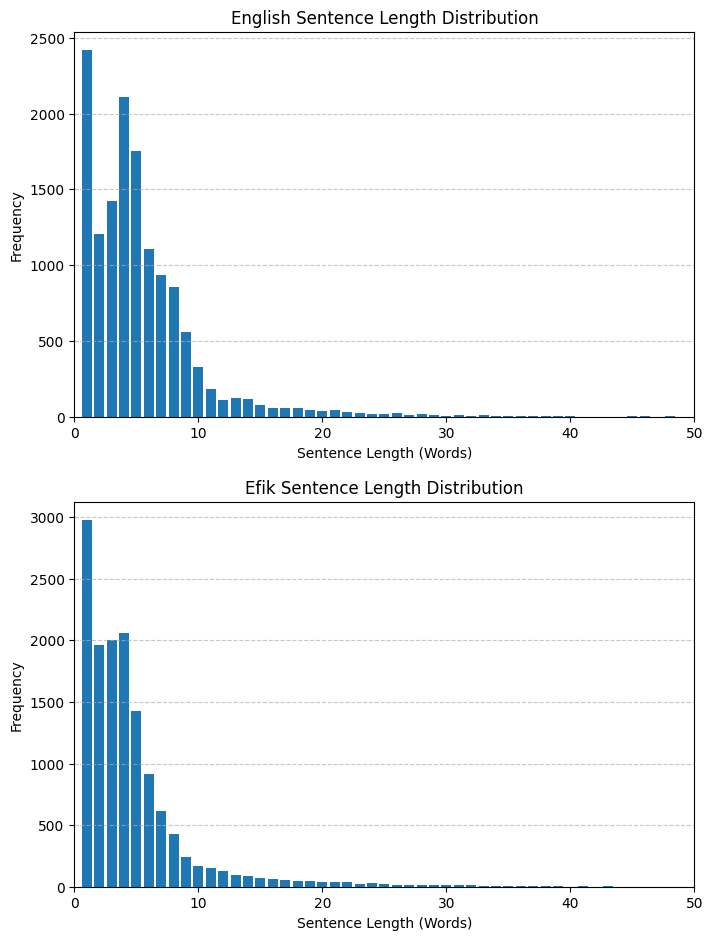}
  \caption{Sentence length distributions for English and Efik.}
  \label{fig:sentence-length}
\end{figure}
 
In summary, the corpus exhibits a wide variety of sentence lengths, which is important for training a translation model that must handle both short conversational utterances and longer, structurally complex sentences. Short sentences typically include greetings, responses, exclamations, and simple conversational turns, while longer sentences commonly originate from educational, descriptive, or religious content.

\subsubsection{Vocabulary Frequency}
\begin{table}[ht]
\centering
\begin{tabular}{l r | l r}
\hline
\textbf{English} & \textbf{Count} & \textbf{Efik} & \textbf{Count} \\
\hline
house & 298 & ke & 3,457 \\
good & 236 & enye & 2,094 \\
want & 220 & mi & 1,407 \\
thing & 211 & oro & 1,087 \\
money & 211 & ye & 830 \\
one & 202 & mme & 791 \\
like & 176 & ama & 765 \\
food & 163 & mm\d{o} & 762 \\
person & 162 & edi & 738 \\
child & 158 & owo & 698 \\
\hline
\end{tabular}
\caption{Top 10 most frequent words in English and Efik along with their counts.}
\label{tab:vocab-frequency}
\end{table}

From Table ~\ref{tab:vocab-frequency}, it can be seen that certain Efik function words appear with very high frequency compared to English, reflecting the language’s morphological and syntactic patterns. Overall, the distribution underscores common challenges in low-resource, morphologically rich languages, including sparse lexical diversity, high repetition of grammatical markers, and increased rates of out-of-vocabulary tokens, which motivates the use of subword tokenization and larger, more diverse corpora to improve translation quality.

\subsection{Model Training}
\subsubsection{Training Setup}
We adopted a transfer-learning approach by fine-tuning two massively multilingual models: mT5-small  \cite{Xue2021mT5}, a text-to-text Transformer pre-trained on 101 languages, and NLLB-200 \cite{NLLB2022}, a sequence-to-sequence model covering over 200 languages. Transfer learning has been shown to be more effective for low-resource languages than training models from scratch \cite{Downey2024TargetedMA}. 

Both models were selected for their efficiency, generalizability, and suitability for multilingual adaptation. Training was conducted using the Hugging Face Transformers library with structured optimization and hyperparameter settings tuned for low-resource English–Efik translation.

\begin{table}[ht]
\centering
\begin{tabular}{l c c}
\hline
\textbf{Hyperparameter} & \textbf{mT5} & \textbf{NLLB} \\
\hline
Learning rate & 1e-4 & 2e-5 \\
Train batch size per device & 8 & 2 \\
Eval batch size per device & 8 & 2 \\
Gradient accumulation steps & 1 & 2 \\
Number of epochs & 20 & 6 \\
Warmup ratio & - & 0.1 \\
Weight decay & 0.01 & - \\
FP16 & - & True \\
\hline
\end{tabular}
\caption{Final hyperparameters used for mT5 and NLLB models.}
\label{tab:final-hyperparameters}
\end{table}


\subsection{Evaluation Methodology}
Translation quality was assessed using both automatic metrics and qualitative error analysis.

\subsubsection{Automatic Evaluation }
We computed BLEU \cite{Papineni2002BLEU} and chrF \cite{Popovi´c2015chrF} using the Hugging Face Evaluate library. BLEU measures n-gram overlap, while chrF, a character n-gram–based metric, captures morphological and orthographic patterns, which is particularly important for languages with tone and inflectional complexity.

Table~\ref{tab:mt-results} summarizes the BLEU and chrF scores for zero-shot and fine-tuned mT5 and NLLB models on English--Efik translation in both directions. The gray-shaded cells indicate the best-performing results for each metric and direction. Both models perform poorly in zero-shot mode, with mT5 largely unable to generate meaningful translations and NLLB producing semantically weak outputs. Fine-tuning substantially improves performance: mT5-FT achieves moderate gains, while NLLB-FT attains the highest scores across all directions and metrics, demonstrating superior handling of semantic meaning and fluency.

\begin{table}[ht]
\centering
\begin{tabular}{l | r r | r r}
\hline
\textbf{Model} 
& \multicolumn{2}{c|}{\textbf{BLEU}} 
& \multicolumn{2}{c}{\textbf{chrF}} \\
 & en--ef & ef--en & en--ef & ef--en \\
\hline

\multicolumn{5}{l}{\textbf{Zero-shot}} \\
mT5-small        & 0.01 & 0.01 & 2.61  & 3.15 \\
\rowcolor{lightgray}
NLLB-200   & 0.11 & 0.22 & 10.95 & 15.16 \\
\hline

\multicolumn{5}{l}{\textbf{Encoder--Decoder}} \\
mT5-small FT     & 14.32 & 18.87 & 35.55 & 36.73 \\
\rowcolor{lightgray}
NLLB-200 FT    & 26.64 & 31.21 & 51.04 & 47.92 \\
\hline
\end{tabular}

\caption{BLEU and chrF scores for zero-shot and fine-tuned mT5 and NLLB models on English--Efik translation.}
\label{tab:mt-results}
\end{table}

\begin{table*}[t]
\centering
\small
\renewcommand{\arraystretch}{1.2}
\begin{tabular}{p{0.20\textwidth} p{0.20\textwidth} p{0.20\textwidth} p{0.20\textwidth}}
\hline
\textbf{Source (En)} & \textbf{Reference (Ef)} & \textbf{mT5-small FT Output} & \textbf{NLLB-200 FT Output} \\
\hline
My mother is very sick and she is in the hospital. &
Eka mi \d{o}d\d{o}\~n\d{o} eti eti ndien enye odu ke uf\d{o}kib\d{o}k. &
Eka mi \textcolor{red}{\d{o}s\d{o}\~n\d{o}}
\textcolor{red}{[\textit{eti eti omitted}]}
ndien enye \textcolor{red}{\d{o}di\d{o}\~n\d{o}} ke uf\d{o}kib\d{o}k. &
Eka mi \d{o}d\d{o}\~n\d{o} eti eti ndien enye odu ke uf\d{o}kib\d{o}k. \\
\hline

Are you carrying that umbrella for the rain or just to look fancy? &
Ndi akama ufukeyo ke ntak ed\d{i}m mme onyu\~n akama ke asian? &
Ndi afo \textcolor{red}{ebet} ufukeyo oro \textcolor{red}{\d{o}n\d{o}} ed\textcolor{red}{i}m mme \textcolor{red}{\~nkp\d{o} etiede}? &
Ndi emen ufukeyo oro ke ntak ed\textcolor{red}{i}m mme \textcolor{red}{ndikut \d{o}kp\d{o}s\d{o}\~n \~nkp\d{o}}? \\
\hline

\end{tabular}
\caption{Qualitative comparison of English to Efik translations produced by mT5 and NLLB.}
\label{tab:qualitative-efik}
\end{table*}

\begin{table*}[t]
\centering
\small
\renewcommand{\arraystretch}{1.2}
\begin{tabular}{p{0.20\textwidth} p{0.20\textwidth} p{0.20\textwidth} p{0.20\textwidth}}
\hline
\textbf{Source (Ef)} & \textbf{Reference (En)} & \textbf{mT5-small FT Output} & \textbf{NLLB-200 FT Output} \\
\hline
Udia usenub\d{o}k amakebe, mm\d{o} ew\d{o}r\d{o} anwa ndinwam ekpri eyeneka esie ndimen akpasa nya inwa\~n mbakara kiet nka efe urua. &
After breakfast, they went outside to help his younger brother carry a basket of tomatoes to the stall. &
\textcolor{red}{The morning went down}, they went to help his little sister pick up a basket of \textcolor{red}{yams} to the \textcolor{red}{market}. &
After breakfast, they set off to help his younger brother carry a basket of tomatoes to the stall. \\
\hline

William Otu ekedi akwa owo ke akamba usiakifia aran is\d{o}\~n ke Lagos. &
William Otu was a seasoned manager at a big oil company in Lagos. &
William \textcolor{red}{group was a very heavy trader in the mining companies}. &
William Otu was a senior official in the Lagos oil company. \\
\hline

\end{tabular}
\caption{Qualitative comparison of Efik to English translations produced by mT5 and NLLB.}
\label{tab:qualitative-english}
\end{table*}

\subsubsection{Qualitative Error Analysis}
We conducted a qualitative error analysis on a sample of test sentences for both mT5-FT and NLLB-FT models. Each translation was carefully inspected for substitutions, omissions, and partially correct words. Errors are highlighted in red and omitted content is indicated using italicized red brackets. This analysis provides additional insights into the strengths and weaknesses of each model, complementing the automatic evaluation metrics (Table~\ref{tab:mt-results}).

Table~\ref{tab:qualitative-efik} presents selected examples of English-to-Efik translations. In the first example, the mT5 model omits the repeated phrase \textit{eti eti}, while NLLB correctly preserves it. In the second example, NLLB produces a more fluent and semantically accurate translation than mT5, which fails from the beginning by introducing spurious content and incorrect word choices. In the English-to-Efik examples, while NLLB preserves most of the meaning, it does miss some contextual nuances at the end of the sentence (e.g., translating “fancy”).

Table~\ref{tab:qualitative-english} presents selected Efik-to-English translations. NLLB maintains the overall semantic meaning of the source sentences, even when the translation is not strictly word-for-word, which is typical and desirable in natural translation. In contrast, mT5 often produces incorrect lexical choices, alters the meaning of domain-specific terms, and fails to correctly translate named entities, such as "Otu". 

Overall, NLLB demonstrates higher fidelity, fluency and robust handling of culturally specific content compared to mT5 in Efik-to-English translation.

\section{Results and Discussion}
Translating between English and Efik remains challenging despite the use of multilingual transfer-learning models. Both mT5 and NLLB perform well on high-frequency words and simple sentences, but they often struggle with idiomatic expressions, morphological inflections, and longer or more complex structures.

Qualitative analysis reveals that mT5 frequently produces incorrect lexical choices, alters domain-specific meanings and sometimes fails to translate named entities correctly. NLLB, in contrast, generally preserves the overall semantic meaning, even when translations are not strictly word-for-word, demonstrating higher fidelity, fluency, and better handling of culturally specific content.

These limitations are partly due to the scarcity of parallel data, the absence of direct Efik equivalents for certain English terms and less commonly used vocabulary. Overall, the results highlight the potential of current multilingual fine-tuning approaches for low-resource languages while emphasizing the need for additional data and broader domain coverage to improve translation quality and coverage.

\section{Conclusion}

This study presents fine-tuned multilingual translation systems for English–Efik, using both mT5-small and NLLB-200 models on a manually curated parallel corpus of 13,865 sentence pairs. The results demonstrate that, even with limited data, NLLB produces translations that better preserve semantic meaning, handle culturally specific content, and maintain fluency compared to mT5, which struggles with lexical accuracy and named entities. Both models, however, face challenges with idiomatic expressions, complex sentence structures, and less common or older vocabulary, reflecting the intrinsic difficulties of low-resource language translation.

Our work contributes a community-curated English–Efik dataset and a reproducible pipeline for fine-tuning multilingual models, providing a foundation for future NLP research on underrepresented African languages. The findings highlight the importance of increasing the volume and diversity of parallel data to improve translation fidelity and coverage. Techniques such as back-translation, data augmentation, and crowdsourced annotation could help bridge current gaps in linguistic representation.

Finally, this study highlights the potential of large and specialized multilingual translation models, as demonstrated by mT5 and NLLB-200, for bidirectional English–Efik machine translation. By combining high-quality parallel data with effective full-parameter fine-tuning, future work can further improve both automatic evaluation metrics and human-perceived translation quality. Overall, this research represents a step toward more inclusive NLP, supporting the development of robust and linguistically informed translation systems for Efik and other low-resource African languages.

\section{Limitations}
Despite promising results, our study has several limitations. The dataset size (13.8K sentence pairs) restricts exposure to idiomatic expressions and domain-specific language. Orthographic variation, particularly in tone marking and dialectal differences within Efik, may reduce translation consistency. While BLEU and chrF \cite{Papineni2002BLEU, Popovi´c2015chrF} provide useful quantitative insights, they do not fully capture human judgments of fluency or cultural appropriateness \cite{Kreutzer2022QualityAtAGlance}. Additionally, our human evaluation was limited in scale due to resource constraints. Future work should expand the dataset, apply phonological normalization, and involve community-driven annotation to better capture linguistic diversity and cultural nuances \cite{Joshi2020TheState}.

\section{Acknowledgments}
We would like to thank Prof.\ Eyo Mensah, Dr.\ David Adelani, Dr.\ Orok Duke, Dr.\ Enoima Umoh, Gideon George, Abraham Owodunni, Steven Kolawole, Luel Hagos and Saheed Azeez for their invaluable feedback and support throughout this work. We also extend our sincere appreciation to Archibong Archibong, Andrew Nkpanam, Pst.\ Praize Aye, Mr.\ Archibong Effiong, Mrs.\ Edung, Elder Edem Mensah, Mrs.\ Ansa Edem, Mr.\ Felix Effiom, Sylvia Edem Mensah, and other native Efik speakers, linguists, and community contributors for their translations and validation efforts. Their linguistic and cultural expertise made this project possible.

We further acknowledge Chief E.\ U.\ Aye for his foundational contribution to Efik language documentation through \textit{A Learner's Dictionary of the Efik Language}, which served as an important reference during dataset development.

We also acknowledge the Africa NLP open-source community, particularly Masakhane, for fostering a collaborative research environment that enabled reproducible research. Finally, we thank the Machine Learning Collective, a community of machine learning researchers, for their support and guidance.

\nocite{*}  
\bibliography{custom}

\appendix



\end{document}